\documentclass[letterpaper, 10 pt, conference]{ieeeconf}  

\IEEEoverridecommandlockouts                              

\usepackage{stfloats} 

\usepackage{amssymb,mathtools}
\usepackage{bm}
\usepackage{graphicx}
\usepackage{hyperref}
\usepackage{cite}

\usepackage{stfloats} 

\usepackage{tikz}
\usetikzlibrary{arrows.meta,calc,positioning}
\usepackage{pgfplots}
\pgfplotsset{compat=1.18}

\newtheorem{definition}{Definition}

\usepackage{graphics} 
\usepackage{epsfig} 
\usepackage{mathptmx} 
\usepackage{times} 
\usepackage{algorithm}
\usepackage{cite}
\usepackage{algpseudocode}
\usepackage{multirow}
\usepackage{glossaries}
\usepackage{CJK}
\usepackage{booktabs}
\setacronymstyle{long-short}
\usepackage[dvipsnames]{xcolor}
\usepackage{hyperref}
\usepackage{enumerate}
\usepackage{microtype} 
\DeclareUnicodeCharacter{2212}{-}
\newcommand{\Tau}{\mathrm{T}}

\title{\LARGE \bf Safety-Constrained Model Predictive Control for an Omnidirectional Walking Assistive Robot Using Control Barrier Function}

\author{Andrea Fortuna$^{1,2}$, Marta Lorenzini$^{1}$, Elisa Motta$^{1}$, Alberto Ranavolo$^{3}$, Elena De Momi$^{2}$, Arash Ajoudani$^{1}$
\thanks{1- Human-Robot Interfaces and Interaction Laboratory, Istituto Italiano di Tecnologia, Genoa, Italy 2- Department of Electronics, Information and Bioengineering, Politecnico di Milano, Milan, Italy. 3- Dep. of Occupational and Environmental Medicine, Epidemiology and Hygiene, INAIL, Rome, Italy. Corresponding author's email: {\tt\small andrea.fortuna@iit.it}}
\thanks{This work was supported by INAIL 
under the BRIC 2025 programme, project PROMETEO, ID23.} \vspace{-0.8cm} 
}

\begin{document}

\maketitle
\thispagestyle{empty}
\pagestyle{empty}

\begin{abstract}
Providing safe and effective mobility assistance plays a crucial role in restoring independence and enhancing the quality of life for individuals with motor impairments.
In this context, robotic walking assistive devices have recently emerged as promising solutions to provide physically compliant interaction while ensuring user safety and support. This paper presents a novel control framework for an omnidirectional Walking Assistive Robot (I-WANDER) that integrates a Control Barrier Function (CBF) formulation into a Model Predictive Control (MPC) scheme to explicitly enforce collision-avoidance safety constraints while optimizing for energy efficiency and smooth human–robot collaboration. The method was experimentally evaluated with 12 healthy participants performing two different walking tasks using both the proposed CBF-based MPC controller (CB-MPC) and a variable admittance controller (AC). The first task involved structured navigation through a U-shaped corridor, whereas the second consisted of a single-obstacle avoidance task performed blindfolded to ensure the obstacle was unexpected. Comparative results show that the CB-MPC architecture significantly reduces energy consumption and mechanical work ($p < 0.01$) without compromising motion smoothness, while also decreasing the number of obstacle collisions. Overall, the findings highlight the potential of the proposed control architecture to enhance both safety and efficiency in robotic walking assistance.
\end{abstract}

\section{INTRODUCTION}
\label{sec: introduction}
Gait dysfunction represents one of the most common causes of disability in Europe, with approximately five million citizens depending on a wheelchair. The progressive ageing of the global population further exacerbates this condition, as chronic diseases associated with ageing lead to an increasing prevalence of gait impairments and mobility-related disabilities. These limitations significantly reduce individuals’ ability to perform activities of daily living (ADLs) and to actively participate in social and professional life \cite{Tadeusz_review}.

To restore, support, and preserve gait function, 
several mechanical assistive devices have been developed, including walkers and canes. However, traditional assistive tools present several limitations, including the need for sufficient user-generated force to maneuver the device, limited adaptability to human motion, and insufficient support for user stability.
In recent years, robotic-assisted devices have emerged as a promising alternative to conventional ambulation tools. Most systems are designed as robotic canes or smart walkers composed of a mobile base and a handle interface \cite{xing2021admittance, ding2022intelligent, itadera2019predictive, itadera2022admittance}. Despite their technological advancements, these platforms typically rely on user-applied forces on the handlebars to generate motion commands, precluding hands-free operation. Moreover, they do not physically sustain the user and therefore cannot prevent vertical falls. 

To address these shortcomings, reduced-dimension robotic platforms that can physically support the user at the trunk level have recently been proposed. The mobile robotic balance assistant (MRBA) introduced in \cite{li2023mobile} implements a user-following strategy that tracks the user’s Center of Mass (CoM) relative to the robot. When the robot-CoM distance exceeds a predefined threshold, the robot moves toward the user. 
Similarly, \cite{mun2014design, mun2015development, aguirre2019high, aguirre2021omnidirectional} proposed an omnidirectional assistive platform equipped with an admittance controller (AC), which ensures smoother human-robot interaction compared to user-following approaches. 
A force/torque (FT) sensor is coupled to the patient to measure interaction forces, which are then converted into desired velocities through an admittance model.

Nevertheless, individuals with gait disabilities often present not only sensorimotor deficits, reduced muscle strength, impaired coordination, 
and limited effort capacity, 
but also varying degrees of cognitive impairment, pain, and movement-avoidance strategies \cite{Tadeusz_review}. These factors can limit their ability to react promptly and adapt to unexpected situations. Therefore, the control strategy of assistive devices must explicitly account for constrained operational scenarios and ensure safe interaction with surrounding obstacles and humans. Notably, none of the aforementioned works explicitly addresses collision avoidance in trunk-support walking assistive devices. 

In general, to ensure safe interaction among robots, surrounding obstacles, and humans, indirect force-control strategies such as impedance control are widely adopted \cite{Johansson_1994}. Nevertheless, pure impedance-based strategies do not explicitly enforce state constraints and may therefore be insufficient to guarantee collision avoidance in complex environments \cite{Ducaju_2025}. In recent years, Control Barrier Functions (CBFs) have been introduced to strengthen obstacle-avoidance capabilities by providing formal guarantees of safe state sets. If the system starts from a collision-free configuration, CBFs ensure that it will not enter unsafe regions. Moreover, CBF-based formulations allow 
the direct translation of environmental safety constraints into control input constraints.
Most existing works implement CBFs within one-step Quadratic Programming (QP) problems \cite{Landi_2019, Ducaju_2022}, thus limiting their ability to adapt control actions in a predictive manner. In \cite{Ducaju_2025}, CBFs are integrated into a Model Predictive Control (MPC) formulation for impedance control in physical Human Robot Interaction (pHRI) scenarios. Nevertheless, the reference trajectory is predefined, and human interaction mainly acts as a corrective input around this nominal trajectory. Thus, the robot motion remains largely constrained by an a priori path. The complexity of pHRI in walking assistive devices, where human intention directly determines motion and the robot must operate in a fully shared and dynamic environment, exceeds the scenarios considered in such formulations.

This study focuses on enabling proactive collision avoidance by embedding CBFs within an MPC framework. The resulting strategy ensures safe ambulation with the I-WANDER walking assistive device, 
avoiding obstacles while minimizing human-robot energy exchange and robot oscillations. 


The rest of the paper is organized as follows. In Sec. \ref{sec: Hardware overview} and Sec. \ref{sec: Optimal control}, an overview of I-WANDER and its control framework are provided, respectively. Sec. \ref{sec: Experiments} explains the experimental protocol and the analysis conducted to evaluate the controller. In Sec. \ref{sec: results}, the results are presented and discussed in Sec. \ref{sec: discussion}. Sec. \ref{sec: conclusions} draws the conclusions.

\section{Hardware Overview}
\begin{figure*}[b]
    \centering
    \includegraphics[trim= 0cm 4cm 0cm 6.2cm,clip,width=1.0\linewidth]{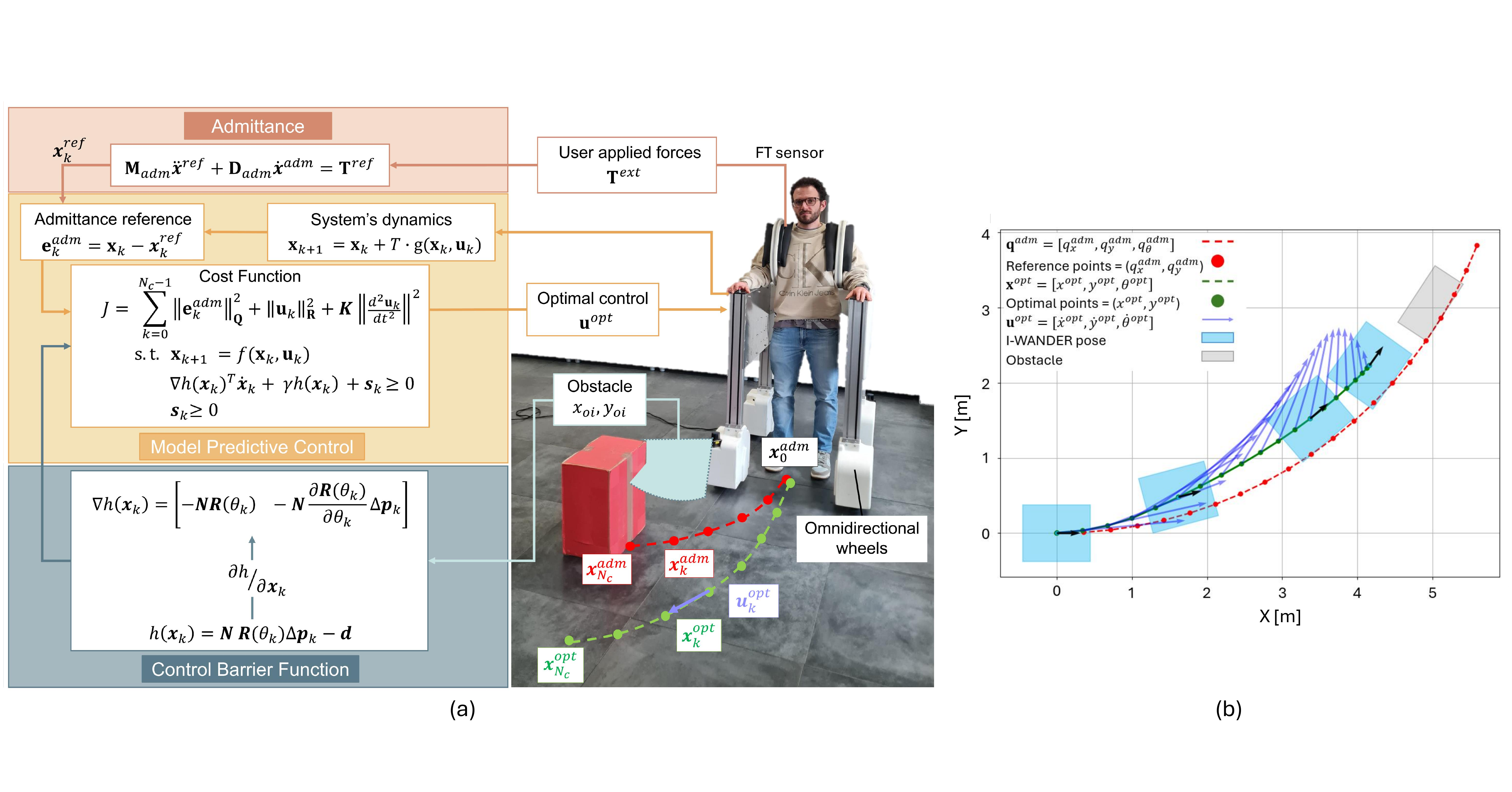}
    \vspace{-0.5cm}
    \caption{(a) Overview of the proposed control schema, including: I-WANDER assistive device; Model Predictive Control (yellow block); Control Barrier Function (grey block), and Admittance Control (orange block). (b) Overview of the effect of CBF integration within the MPC framework: by enforcing CBF constraints over the entire prediction horizon, the control input $\mathbf{u}^{opt} = [\dot{x}^{opt}, \; \dot{y}^{opt}, \;\dot{\theta}^{opt}]$ is predictively adjusted at each step to simultaneously ensure obstacle avoidance and minimize jerk.}
    \label{fig:control_framework_placeholder}
    \vspace{-0.4cm}
\end{figure*}

\vspace{-0.1cm}
\label{sec: Hardware overview}
The I-WANDER platform, depicted in Fig.~\ref{fig:control_framework_placeholder}a, is equipped with four omnidirectional wheels mounted at the corners of its base, enabling planar motion with three degrees of freedom, namely longitudinal translation, lateral translation, and on-the-spot rotation. 
The mechanical structure of the platform encloses the user, enhancing awareness of the robot’s footprint and improving visual perception of its overall dimensions, thereby ensuring intuitive maneuverability. A trunk support interface (TSI) establishes a rigid mechanical connection between the user and the robotic device, maintaining proximity between the human Center of Mass (CoM) and that of the device. Under this configuration, the human–robot system can be approximated as a single body with coincident positions, thereby reducing the user's perceived inertia during motion. Interaction forces and torques applied by the user are measured through a Bota LaxOne six-axis force/torque (FT) sensor integrated at the center of the TSI.

\subsection{LiDAR system}
\label{sec: lidar overview}
Two 2D Hokuyo LiDAR sensors are mounted on the front-right and rear-left sides of the I-WANDER platform at a height of 30 cm above the ground. These sensors detect obstacle point clouds within their respective fields of view. The maximum sensing range is set to 5 m, and each LiDAR operates over an angular range from $-90^{\circ}$ to $+180^{\circ}$ at a sampling frequency of 40 Hz. The combined configuration of the front and rear sensors enables obstacle detection along the entire perimeter of the I-WANDER. The point clouds acquired by the two LiDARs are transformed into a common local reference frame located at the center of the mobile base footprint. To ensure adequate coverage of all surrounding obstacles, the selected points from the point cloud are filtered such that adjacent points are spaced by at least 0.10 m. To limit computational complexity within the control framework, only a subset of obstacle points is considered in the optimization problem. Specifically, a maximum of $N_{obs} = 20$ points corresponding to the closest obstacles to the robot are selected for further processing.

\section{Optimal Control enhanced with safe obstacle avoidance}
\label{sec: Optimal control}
In this section, we provide an overview of the control system developed to ensure safe obstacle avoidance while guaranteeing smoothness and energy-efficient ambulation with the I-WANDER. Similar to \cite{itadera2022admittance}, an AC is employed to enable intuitive human–robot interaction by converting the user-applied forces into motion references. The AC (orange block in Fig. \ref{fig:control_framework_placeholder}a) generates the reference state $\mathbf{x}^{ref} = [x, y, \theta]^\top$, representing the pose of the coupled human–robot system relative to the robot’s pose at the initial prediction step. 
Subsequently, an MPC scheme (yellow block in Fig. \ref{fig:control_framework_placeholder}a) computes the optimal control inputs by tracking the reference trajectory provided by the AC while simultaneously enforcing the safety constraints encoded through the CBF (gray block in Fig. \ref{fig:control_framework_placeholder}a). The effect of the CBF is illustrated in Fig.\ref{fig:control_framework_placeholder}b), which shows one MPC solution over the prediction horizon. In this example, the reference trajectory generated by the AC (red points) would result in a collision with a detected obstacle. By enforcing the safety constraint, the CBF yields a safe optimized trajectory (green points), allowing the system to avoid the obstacle in advance.

\subsection{Admittance Control}
\label{sec: Admittance control}
The admittance dynamics are modeled as
\begin{equation}\label{Eq: admittance control}
\mathbf{M}_{adm}\mathbf{\ddot{x}}^{ref} + \mathbf{D}_{adm}\mathbf{\dot{x}}^{ref} = \mathbf{\Tau}^{ext},
\end{equation}
where $\mathbf{\ddot{x}}^{ref}$ and $\mathbf{\dot{x}}^{ref}$ denote the desired accelerations and velocities, respectively. The external wrench $\mathbf{\Tau}^{ext} = [F_x, F_y, \tau_z]^\top$ corresponds to the forces and torque measured by the six-axis FT sensor and appropriately mapped into the robot frame. The matrices $\mathbf{M}_{adm} = \mathrm{diag}(M_x, M_y, M_\theta)$ and $\mathbf{D}_{adm} = \mathrm{diag}(D_x, D_y, D_\theta)$ represent the positive definite diagonal matrices of virtual mass and damping defined in Cartesian coordinates. To construct the reference trajectory over the prediction horizon, ${\mathbf{X}^{ref}} = \{ \mathbf{x}^{ref}_{0}, \ldots, \mathbf{x}^{ref}_{N_c} \}$, the admittance dynamics in Eq. \eqref{Eq: admittance control} are numerically integrated for $N_c$ steps using a fourth-order Runge–Kutta (RK4) scheme, assuming constant user-applied interaction forces. 

\subsection{Problem formulation}
The considered scenario can be interpreted as a human-following robotic system, where the control objective is to ensure that the I-WANDER behaves compliantly with respect to the motion of the coupled user, reproducing the user’s intended movement as accurately as possible. Accordingly, the control task is formulated as a reference-trajectory-tracking problem within a receding-horizon optimization framework. Given that I-WANDER is a four-wheeled omnidirectional mobile platform, the following definitions are introduced:

\begin{itemize}
    \item $\mathbf{x} = [x, y, \theta]^\top$: state vector of the coupled human–robot system, where $(x, y)$ represent the planar position expressed in the robot frame at the initial prediction step, and $\theta$ denotes the robot orientation;

    \item $\mathbf{u} = [\dot{x}, \dot{y}, \dot{\theta}]^\top$: control input vector, where $\dot{x}$ and $\dot{y}$ correspond to the linear velocities along the local $x$ and $y$ axes of the robot, and $\dot{\theta}$ is the angular velocity about the vertical axis;

    \item The system dynamics are described by a discrete-time nonlinear model with sampling time $T$ and time step $k$:
    \begin{equation}
    \label{eq:system_dynamics}
    \mathbf{x}_{k+1} = f(\mathbf{x}_k, \mathbf{u}_k) 
    = \mathbf{x}_k + T \, g(\mathbf{x}_k, \mathbf{u}_k),
    \end{equation}
    where $\mathbf{x} = [{x}, {y}, {\theta}]^\top$ is the state computed at each horizon step and the nonlinear mapping $g(\cdot)$ is defined as
    \[
    g(\mathbf{x}_k, \mathbf{u}_k) =
    \begin{bmatrix}
    \cos(\theta_k)\dot{x}_k - \sin(\theta_k)\dot{y}_k \\
    \sin(\theta_k)\dot{x}_k + \cos(\theta_k)\dot{y}_k \\
    \dot{\theta}_k
    \end{bmatrix}.
    \]
\end{itemize}
The function $g(\mathbf{x}_k, \mathbf{u}_k)$ computes the state trajectory as a function of the control input $\mathbf{u} = [\dot{x}, \dot{y}, \dot{\theta}]^\top$, expressed in the robot 
frame at the initial prediction step ($N_c = 0$). Using this formulation, both the reference trajectory $\mathbf{x}^{ref} = [x, y, \theta]^\top$ and the predicted trajectory $\mathbf{x}$ are represented within the same coordinate frame. This alignment ensures consistency in the objective function formulation, which is defined as follows. 
\begin{equation}
J = 
\sum_{k=0}^{N_c-1} 
\!\!\
\|\mathbf{e}_k^{ref}\|_{\mathbf{Q}}^{2} 
+ \|\mathbf{u}_k\|_{\mathbf{R}}^{2} 
+ \mathbf{K}\!\left\|\tfrac{d^{2}\mathbf{u}_k}{dt^{2}}\right\|^{2}
\
\label{eq:mpc_cost_compact}
\end{equation}
where $\mathbf{e}_k^{ref} = \mathbf{x}_k - \mathbf{x}_k^{ref}$ is the error between the expected position calculated by means of the dynamic model and the reference created by the AC. The term $\mathbf{u}_k$ penalizes high values of the control input, while the last term in the cost function minimizes the oscillation of the mobile robot. $\mathbf{Q}, \mathbf{R} \in \mathbb{R}^{3 \times 3}$ are positive definite weighting matrices penalizing deviations from the reference trajectory in state and control space, whereas $\mathbf{K} \in \mathbb{R}^{3 \times 3}$ is a diagonal weight matrix penalizing the jerk term.

Minimizing the quadratic cost function $J(\mathbf{x}, \mathbf{u})$ enables accurate tracking of the reference trajectory generated from the user’s intended direction of motion, while ensuring smooth, compliant human-robot interaction. 


\subsection{Control Barrier Function Formulation}
\label{sec:cbf formulation}
The CBF formulation is embedded within the Optimal Control Problem (OCP), which enforces safety constraints to guarantee obstacle avoidance during operation.
Consider the control-affine nonlinear system
\begin{equation}
\mathbf{\dot{x}} = f(\mathbf{x}) + g(\mathbf{x})\mathbf{u},
\label{eq:affine_system}
\end{equation}
where $\mathbf{x} \in \mathbb{R}^n$ denotes the state vector and 
$\mathbf{u} \in \mathbb{R}^m$ the control input. Let $h(\mathbf{x})$ be a continuously differentiable function encoding safety through the forward-invariant set
\[
\mathcal{C} = \{ \mathbf{x} \in \mathbb{R}^n \mid h(\mathbf{x}) \ge 0 \}.
\]
While safety is defined in terms of the state $\mathbf{x}$, the mobile base is controlled via the input $\mathbf{u}$. Therefore, the state constraint must be translated into a constraint on $\mathbf{u}$.

To this end, consider the differential inequality, with $\gamma > 0$
\begin{equation}
\dot{h}(\mathbf{x}) \ge -\gamma h(\mathbf{x}).
\label{eq:cbf_diff_ineq}
\end{equation}
The equality case $\dot{h} = -\gamma h$ yields the solution $h(\mathbf{x}) = h_0 e^{-\gamma \mathbf{x}}$, which guarantees $h(\mathbf{x}) > 0$ for all $\mathbf{x}$ provided $h_0 > 0$. 
Relaxing the equality to the inequality in Eq. \eqref{eq:cbf_diff_ineq} enlarges the admissible control set while still ensuring forward invariance of $\mathcal{C}$. 
The scalar $\gamma$ regulates the conservativeness of the constraint.

Applying the chain rule to $h(\mathbf{x})$ along the system trajectories yields
\begin{equation}
\dot{h}(\mathbf{x}) = \nabla h(\mathbf{x})^\top f(\mathbf{x}) 
+ \nabla h(\mathbf{x})^\top g(\mathbf{x})\mathbf{u}.
\end{equation}

Defining the Lie derivatives
\begin{equation}
L_f h(\mathbf{x}) = \nabla h(\mathbf{x})^\top f(\mathbf{x}), 
\qquad
L_g h(\mathbf{x}) = \nabla h(\mathbf{x})^\top g(\mathbf{x}),
\label{eq:lie_derivatives}
\end{equation}
we obtain
\begin{equation}
\dot{h}(\mathbf{x}) = L_f h(\mathbf{x}) + L_g h(\mathbf{x})\mathbf{u}.
\label{eq:hdot_lie_form}
\end{equation}

Substituting \eqref{eq:hdot_lie_form} into \eqref{eq:cbf_diff_ineq} results in the input-affine constraint
\begin{equation}
L_f h(\mathbf{x}) + L_g h(\mathbf{x})\mathbf{u} \ge -\gamma h(\mathbf{x}),
\label{eq:cbf_input_constraint}
\end{equation}
which enforces safety directly through the control input.

\begin{definition}
A continuously differentiable function 
$h:\mathbb{R}^n \rightarrow \mathbb{R}$ 
is a Control Barrier Function (CBF) if, 
for all $\mathbf{x} \in \mathbb{R}^n$ and for some $\gamma > 0$, there exists a control input $\mathbf{u}$ such that
\[
L_f h(\mathbf{x}) + L_g h(\mathbf{x})\mathbf{u} \ge -\gamma h(\mathbf{x}).
\]
\end{definition}

In an OCP, safety needs to be ensured without compromising tracking performance. Given a nominal control input $\mathbf{u}_{\mathrm{des}}(t)$ (e.g., generated by MPC scheme), the CBF computes the minimally modified safe input
\begin{IEEEeqnarray}{rCl}
\mathbf{u}^\star(t) &=& 
\arg\min_{u} 
\;\; \|\mathbf{u} - \mathbf{u}_{\mathrm{des}}(t)\|^2
\label{eq:cbf_qp_cost_rew} \\
&& \text{s.t.} \quad 
L_f h(\mathbf{x}) + L_g h(\mathbf{x})\mathbf{u} \ge -\gamma h(\mathbf{x}).
\label{eq:cbf_qp_constraint_rew}
\end{IEEEeqnarray}


\subsection{CBF for the I-WANDER Platform}
\label{sec:cbf a-robot}
In the considered framework, higher-order dynamic terms (e.g., inertia and friction) are neglected, yielding the simplified control-affine model
\begin{equation}
\dot{x} = g(\mathbf{x}_k, \mathbf{u}_k).
\label{eq:simplified_affine}
\end{equation}
with the function $g(\mathbf{x}_k, \mathbf{u}_k)$ defined in Eq. (\ref{eq:system_dynamics})

Consider the coordinates $\mathbf{x}_{oi} = [x_{i}, \; y_{i}]$ of the $ i$-th obstacle, detected by the lidars, and expressed wrt the current pose of the robot in the real world. In the MPC loop the obstacle position $\mathbf{O}_i = [O_{ix}, \; O_{iy}]$ expressed wrt the robot's pose $\mathbf{x}_k$ at the current horizon step $k$ can be expressed as

\begin{equation}
\mathbf{O}_i
=
\mathbf{R}(\theta_k) 
\Delta \mathbf{p}_k 
\label{eq:obstacle_local_rew}
\end{equation}
where
\vspace{-0.1cm}
\begin{equation*}
\mathbf{R}(\theta_k)
=
\begin{bmatrix}
\cos\theta_k & \sin\theta_k \\
-\sin\theta_k & \cos\theta_k
\end{bmatrix}, 
\qquad
\Delta \mathbf{p}_k = \mathbf{x}_{oi} - \mathbf{x}_k.
\end{equation*}

As the reference frame is in the middle of the I-WANDER rectangular footprint, to guarantee that obstacles remain outside the assistive robot (length $L$, width $W$), the following conditions must hold:
\begin{equation}
|O_{ix}| > a, \qquad |O_{iy}| > b,
\end{equation}
where $a=L/2$ and $b=W/2$.
The above geometric conditions define four distinct barrier functions, 
each associated with one face of the 
footprint of the I-WANDER, defined as

\begin{equation}
\mathbf{h}(x_k) =
\mathbf{N} \mathbf{R}(\theta_k) \Delta \mathbf{p}_k - \mathbf{d}
\end{equation}
where
\begin{equation*}
\mathbf{N} =
\begin{bmatrix}
1 & 0 \\
-1 & 0 \\
0 & 1 \\
0 & -1
\end{bmatrix},
\qquad
\mathbf{d} =
\begin{bmatrix}
a \\ a \\ b \\ b
\end{bmatrix}.
\end{equation*}

Each function $h_i(\mathbf{x}_k)$ enforces the obstacle to remain outside the corresponding face of the rectangular robot footprint. 
Safety is guaranteed when
\[
h_i(\mathbf{x}_k) > 0, \quad i = 1,\dots,4.
\]

Therefore, the overall CBF condition is expressed as a system of inequalities composed of the four individual barrier functions. The corresponding gradients with respect to the state vector 
$\mathbf{x}_k$ are given by

\begin{equation}
\nabla \mathbf{h}(\mathbf{x}_k)
=
\begin{bmatrix}
- \mathbf{N} \mathbf{R}(\theta_k) &
- \mathbf{N} \dfrac{\partial \mathbf{R}(\theta_k)}{\partial \theta_k}\,\Delta\mathbf{p}_k
\end{bmatrix},
\end{equation}
where
\begin{equation*}
\dfrac{\partial \mathbf{R}(\theta_k)}{\partial \theta_k}\
=
\begin{bmatrix}
-\sin\theta_k & \cos\theta_k \\
-\cos\theta_k & -\sin\theta_k
\end{bmatrix}.
\end{equation*}

Hence, both the barrier function and its gradient are structured as a set of four constraints, one per robot face, which are selectively enforced during MPC optimization based on the identified active face. Before each MPC iteration, obstacles are expressed in the robot frame and assumed constant over the prediction horizon. 
For each obstacle and horizon step, the closest face of the rectangular footprint is identified as the active face, and only the corresponding barrier function is enforced. Including the constraint (\ref{eq:cbf_input_constraint}) in our OCP yields to the final formulation

\begin{subequations}
\begin{align}
\min_{\mathbf{x},\mathbf{u}} \quad 
& \sum_{k=0}^{N_c-1} {J}(\mathbf{x}_k,\mathbf{u}_k)
\label{eq:ocp_cost} \\
\text{s.t.} \quad & \mathbf{x}_{k+1} = f(\mathbf{x}_k, \mathbf{u}_k) \\ 
&\nabla h(\mathbf{x}_k)^\top \dot{\mathbf{x}}_k 
+ \gamma h(\mathbf{x}_k) + \mathbf{s}_k \ge 0
\label{eq:cbf_slack_final}, 
\\
& \mathbf{s}_k \geq 0, 
\label{eq:slack}
\end{align}

\end{subequations}

where $\gamma$ is a scalar parameter and a nonnegative slack variable $\mathbf{s}_k \ge 0$ is introduced to improve feasibility.


\subsection{Directional-based variable admittance controller}
\label{sec: force-based variable admittance controller}
Variable admittance controller is a well-known technique in literature used to enforce safety in pHRI \cite{VAC, VAC2}. Therefore, a variable AC based on the user-intended direction of motion and the relative distance to the closest obstacle is implemented for comparison. The AC is described by Eq. (\ref{Eq: admittance control}) and, similar to the approach described in \cite{fortuna_correlation}, an indirect force strategy is implemented by modulating the nominal damping value $\mathbf{D}_{adm}$ based on the relative positions of obstacles and the user's current motion direction.

Let $\mathbf{u} = [\dot{x},  \; \dot{y}]^\top$ denote the planar velocity of the system and $\hat{\mathbf{u}} = \mathbf{u}/\|\mathbf{u}\|$ its normalized direction. For each detected obstacle $i$, we compute the relative position unit vector

\begin{equation}
\Delta\mathbf{\hat p_i} = \frac{\Delta \mathbf{p}_i}{\|\Delta \mathbf{p}_i\|}.
\end{equation}

The anisotropic addition of damping along the direction $\Delta\mathbf{\hat p_i}$ is constructed as follows:

\begin{equation}
\Delta \mathbf{D} = \sum_{i=0}^{N_{obs}} \alpha_i \, {\Delta \mathbf{p}}_i \Delta\mathbf{\hat p_i}^\top,
\end{equation}
where $\alpha_i$ determines the obstacle-dependent damping intensity and it is defined as
\begin{equation}
\alpha_i = D_{\max} \, w_d(d_i) \, g_{\text{dir}} ,
\end{equation}
where $D_{\max}$ is the maximum damping increment, $d_i$ is the distance to obstacle $i$, and $g_{\text{dir}}$ is a directional gating term defined as
\begin{equation}
g_{\text{dir}} = \max \left( 0, \hat{\mathbf{u}}^\top \Delta\mathbf{\hat p_i} \right),
\end{equation}
which, by means of the scalar product between the two vectors $\hat{\mathbf{u}}$ and $\Delta\mathbf{\hat p_i}$, ensures that damping increases only when the system is moving toward the obstacle.

The distance-based weighting function is defined as
\begin{equation}
w_d(d_i) =
\begin{cases}
0, & d_i \geq d_0, \\
1, & d_i \leq d_{\min}, \\
\dfrac{d_0 - d_i}{d_0 - d_{\min}}, & \text{otherwise},
\end{cases}
\end{equation}
where $d_0$ denotes the activation distance and $d_{\min}$ the saturation distance.

The resulting damping matrix provided to the admittance controller is therefore
\begin{equation}
\mathbf{D} = \mathbf{D}_{adm} + \Delta \mathbf{D}.
\end{equation}

\begin{figure*}[b]
    \centering
    \includegraphics[trim= 0cm 10cm 0cm 14cm,clip,width=1.0\linewidth]{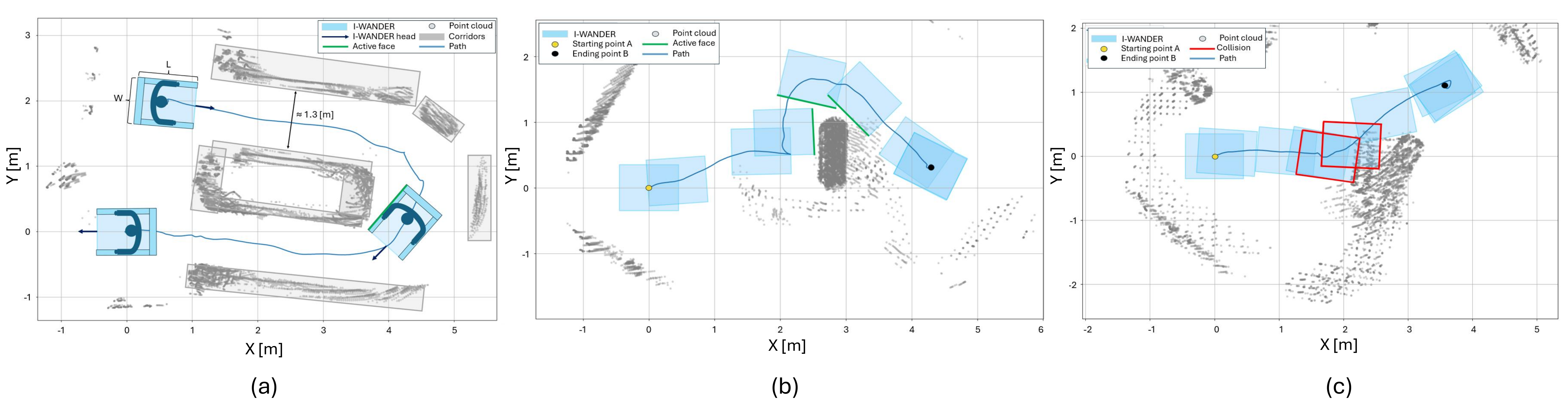}
    \vspace{-0.5cm}
    \caption{Example of a recorded trials showing an overview of the experimental protocol.  Obstacles are detected by the LiDARs sensors as point clouds (gray marks). The I-WANDER footprint, with length $L = 0.9$\,m and width $W = 0.8$\,m, is depicted in light blue. When the user approaches an obstacle, the robot's face closest to the obstacle (highlighted in green) is activated to construct the corresponding Control Barrier Function (CBF). The complete trajectory followed by the user during the analyzed trial is shown in blue. In (a), the \textit{structured navigation} experiment is shown: a set of obstacles (gray rectangles) forms a U-shaped corridor with a width of $\simeq{1.3}$\,m. In (b) and (c), the \textit{unforeseen single-obstacle avoidance} experiment is shown, in the case of CB-MPC and AC, respectively. The obstacle here is transversely aligned with the front face of the I-WANDER.
   }
    
    \label{fig:exp set up}
    \vspace{-0.4cm}
\end{figure*}

\subsection{Controller parameters}
The CB-MPC controller's parameters, reported in Table~\ref{Tab:ExpPar}, were empirically tuned. In particular, the values of admittance matrices $\mathbf{M}_{adm}$ and $\mathbf{D}_{adm}$ were selected to facilitate the user's motion with low physical effort, whereas the oscillations were minimized thanks to the jerk term in the cost function (see Eq. \ref{eq:mpc_cost_compact}). The parameters $N_c = 20$ and $T = 0.1$ [s] were selected to achieve a trade-off between prediction capability and computational cost. This choice corresponds to a prediction horizon of 2 [s] in the OCP, which enables anticipatory adaptation of the control input for collision avoidance. The weighting matrices $\mathbf{Q}$, $\mathbf{R}$, and $\mathbf{K}$ were tuned to ensure accurate trajectory tracking while limiting jerk and preserving smooth human–robot interaction. The parameter $\gamma$ regulates the magnitude of the CBF-based safety constraint. Larger values induce a stronger and earlier corrective action as the system approaches the boundary of the safe set, resulting in a more conservative obstacle-avoidance behavior. Typical values of $\gamma$ lie in the range $[0.2, 1.5]$. In this study, $\gamma = 1$ was selected to balance the allowable proximity to obstacles with the objective of ensuring collision avoidance.

Regarding the AC controller, since it does not rely on an optimization problem, unlike CB-MPC, the nominal values of $\mathbf{M}_{adm}$ and $\mathbf{D}_{adm}$ were calibrated for each subject using Preference-Based Optimization (PBO) to simultaneously minimize physical effort and jerk, as described in \cite{fortuna2024personalizable}. The parameters $d_0, d_{min}$ and $D_{max}$ were selected to maintain a sufficiently large free-motion region around obstacles, in which the additional damping term $\Delta \mathbf{D}$ is inactive, and ensuring adequate space and time to inject additional damping when needed to assist the user in preventing collisions.

\begin{table}[h]
\centering
\caption{Parameters selected for the controllers.}
\label{Tab:ExpPar}
\begin{tabular}{llll}
\toprule
\textbf{CB-MPC} & \textbf{Value}     & \textbf{AC} & \textbf{Value}          \\
\midrule
$\mathbf{M}_{adm}$, $\mathbf{D}_{adm}$  & $\mathrm{diag}(20, 20, 20)$ & $\mathbf{M}_{adm}$, $\mathbf{D}_{adm}$ & PBO \\
 $N_c$ & $20$ & $D_{max}$ & $400$ \\
$\mathbf{Q}$ & $\mathrm{diag}(15, 15, 45)$ & $d_{0}$ & $0.8$ [m] \\
$\mathbf{R}$ & $\mathrm{diag}(25, 25, 50)$ & $d_{min}$ & $0.1$ [m]\\
$\mathbf{K}$ & $\mathrm{diag}(0.1, 0.1, 0.3)$ &  & \\
$T$ & $0.1$ [s] &  &  \\
$\gamma$ & $1$ &  &  \\

\bottomrule \\ 
\multicolumn{3}{l}{}
\end{tabular}
\vspace{-10 mm}
\end{table}

\section{Experiments}
\label{sec: Experiments}
This section presents the experiments conducted to validate the proposed control framework's performance. 

Twelve healthy volunteers, eight males, and four females (age: $29.25 \pm 3.37$ years; mass $71.67 \pm 11.02$ Kg and height $177.42 \pm 8.49$ cm), with no history of walking and balance disability, were recruited for the experiments. 
Prior to the experiments, the TSI height and width were adjusted according to each participant. Then, the subjects were connected to the platform, achieving a rigid coupling without affecting user comfort. 
Two experimental protocols were conducted: a \textit{structured navigation} task and an \textit{unforeseen single-obstacle avoidance} task. In both protocols, the developed controller (CB-MPC) was evaluated against the AC. 


\subsection{Structured navigation experiment}
\label{sec: experimental protocol a}

This task was designed to evaluate the performance of the two controllers under comparable conditions. As shown in Fig.~\ref{fig:exp set up}, the experimental setup reproduced a U-shaped corridor using artificial obstacles. The corridor had a width of $\simeq{1.30}$ m and an overall length of approximately $\simeq{12.0}$ m.
Each participant was instructed to traverse the entire path at a self-selected comfortable pace. The experiment was conducted under the two control configurations CB-MPC and AC. The order of the control conditions was randomized across participants. Four repeated trials were performed for each controller, two in one direction of the corridor and two in the opposite direction, yielding a total of eight per participant. 

\subsection{Unforeseen single-obstacle avoidance experiment}
\label{sec: experimental protocol b}
This task consisted of navigating from a predefined starting point (A) to a target point (B) while walking with I-WANDER. The distance from A to B was $\simeq{5.0}$ m. Subjects were instructed to reach B following the shortest possible path, without prior knowledge of the obstacle configuration. Then, they were blindfolded to perform the task, thus preventing visual perception of the environment and ensuring that obstacle avoidance relied solely on the controller's intervention. For each controller (CB-MPC and AC), four trials were performed under different environmental conditions. One trial was conducted without obstacles. In the remaining three trials, a single obstacle was introduced at different positions relative to the robot: transversely aligned with the front face of the platform (Front), $45^\circ$ to the right of the front face ($45^\circ$ R), and $45^\circ$ to the left ($45^\circ$ L). These configurations were selected to assess the controller’s behavior under distinct collision-risk scenarios. 
The order of the trials was randomized, and subjects were not informed about the presence or location of obstacles before each trial.

\begin{figure*}[b]
   \centering
  \includegraphics[trim= 3cm 14cm 5cm 14cm,clip,width=1.0\linewidth]{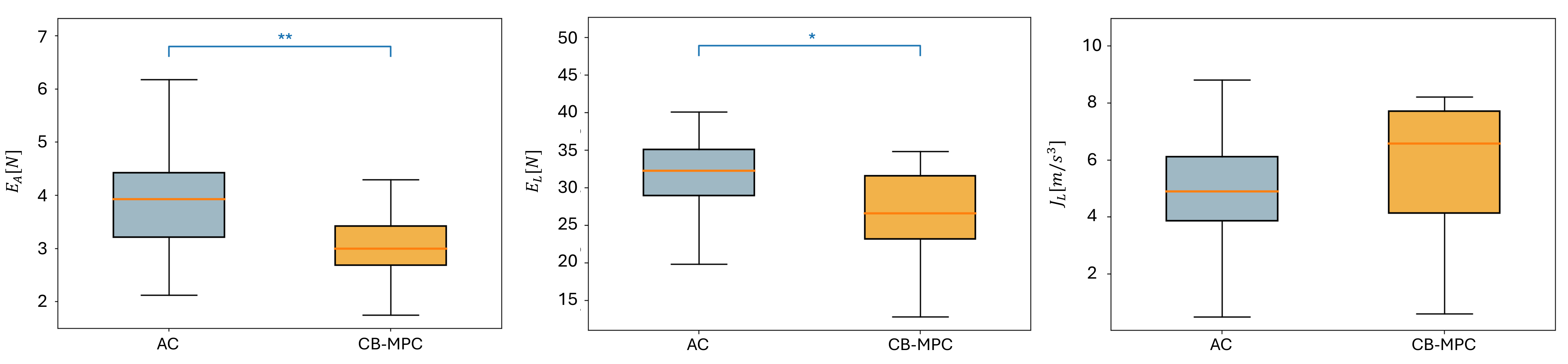}
  \vspace{-0.6cm}
 \caption{Results of the angular energy, linear energy, and jerk for all the subjects in the two experimental conditions (AC, CB-MPC). $**$ and $*$ stand for $p < 0.01$ and $p < 0.05$ respectively.}
\label{fig:results}
\vspace{-0.4cm}
\end{figure*}

\subsection{Experimental analysis}
\label{sec:exp analysis a}

For the evaluation, 
quantitative and qualitative metrics were employed, 
testing the significance of the results. Specifically, an ANOVA with repeated measures and a within-subjects design was conducted to compare 
AC and CB-MPC conditions. For the qualitative metrics, given the reduced data, the Wilcoxon signed-rank test was employed.
\subsubsection{Quantitative metrics}
To evaluate the transparency of the I-WANDER assistance, the required energy per unit distance was used (the lower, the better). The latter comprises the linear energy $E_L$ for forward/backward and lateral movements, and the angular energy $E_A$ for rotational movements.

\[{E}_{L}= \frac{\int_0^s |F|ds}{s}, \quad
{E}_{A}= \frac{\int_0^\theta |\tau_z|d\theta}{\theta},
\]
where $s$ is the total linear path and $\theta$ is the total angular rotation performed during the task's execution.\\
To evaluate the smoothness of the I-WANDER assistance, we used the mean value of the module of the jerk
\begin{equation}
    {J}_{mean}= \frac{\sum_{l=0}^{L}\sqrt{J_x^2(l) + J_y^2(l)}}{L},
\end{equation}
where the jerk in the horizontal directions is defined as \cite{wang2018experimental} 
\[J_x = \frac{d^3 q_x}{dt^3}=\frac{d\ddot{q_x}}{dt}, \quad
J_y = \frac{d^3 q_y}{dt^3}=\frac{d\ddot{q_y}}{dt}.
\]
$l$ iterates the acceleration samples collected from the mobile base, and $L$ is the total number of samples. 

User physical effort was additionally quantified through the mechanical work exchanged between the user and the robot. First, the instantaneous mechanical power generated by the user was computed as
\begin{equation}
    P(t) = F_x(t) u_x(t) + F_y(t) u_y(t),
\end{equation}
where $F_x$ and $F_y$ denote the planar interaction forces expressed in the robot base frame. $u_x$ and $u_y$ represent the planar velocities of the mobile base.
Positive power ($P(t) > 0$) corresponds to propulsive effort, whereas negative power ($P(t) < 0$) represents braking or resisting effort exerted by the user. Accordingly, the cumulative propulsive and braking work were then separately computed as
\begin{equation}
    W^{+} = \frac{\int_0^T P(t)^+ \, dt}{s},
    \qquad
    W^{-} = \frac{\int_0^T P(t)^- \, dt}{s}.
\end{equation}
To enable comparisons across trials with different trajectory lengths, work was normalized by the total travel distance $s$.
Positive work $W^{+} $ can be interpreted as a metric closely related to energy expenditure, as it quantifies the effort required to generate forward motion along the direction of travel. In contrast, negative work $W^{-}$ reflects the force applied to the I-WANDER in opposition to the direction of motion and can therefore be associated with oscillatory behavior during walking.
Finally, the number of collisions with obstacles was evaluated for the two controllers. 

\subsubsection{Qualitative metrics}
At the end of each experimental condition, participants were asked to complete the NASA-TLX\cite{HART1988139} to rate perceived workload for each controller.

\section{Experimental Results}
\label{sec: results}
\subsubsection{Quantitative metrics}
In Fig. \ref{fig:results}, the boxplots of the angular energy, linear energy, and jerk, for all the subjects, are presented in the two tested conditions (AC and CB-MPC), during the \textit{structured navigation} experiment.  
The angular and linear energy were significantly reduced in CB-MPC wrt AC by $14.1\%$ and $27.1\%$, respectively. 
Using the CB-MPC method, no significant changes in jerk with respect to AC were observed. Moreover, the positive work was significantly reduced in CB-MPC wrt AC by $20.93\%$ while the negative work remained almost the same in the two conditions, with a non-significant increase.
In the \textit{unforeseen single-obstacle avoidance} experiment, no statistically significant differences were observed for these metrics.

In Table~\ref{Tab:collision}, the number of collisions is reported for each path configuration. 
During the \textit{structured navigation} task (Corridor), 3 collisions were recorded with the AC controller and 1 with the CB-MPC controller. 
In the \textit{unforeseen single obstacle avoidance} task (the remaining paths), only 1 collision was observed under CB-MPC, whereas 21 collisions were registered with AC. 
Overall, 24 collisions occurred with AC, compared to only 2 with CB-MPC.

\subsubsection{Qualitative metrics}
\begin{figure*}[t]
   \centering
  \includegraphics[trim= 0cm 14cm 0cm 14cm,clip,width=1.0\linewidth]{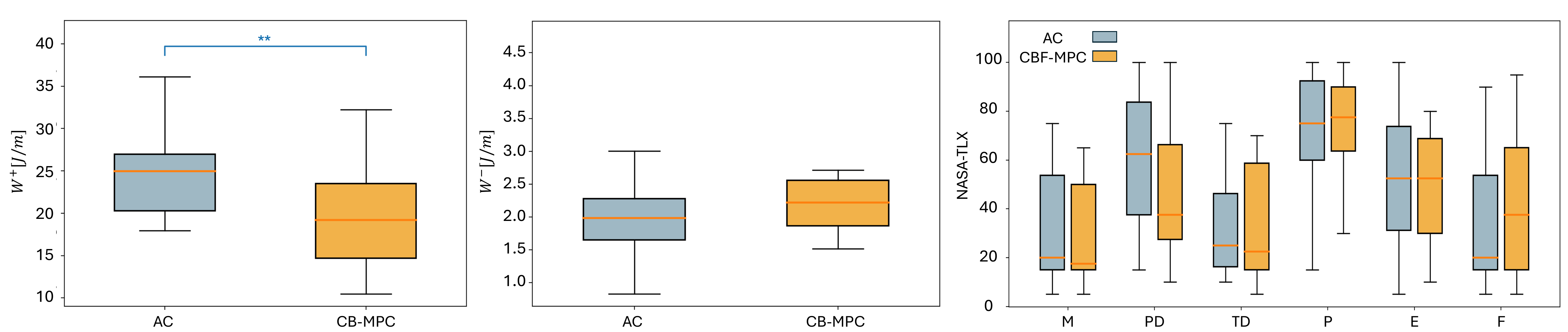}
  \vspace{-0.6cm}
 \caption{Results of the positive and negative work and NASA-TLX questionnaire for all the subjects in the two experimental conditions (AC, CB-MPC). For NASA-TLX, the boxplots for mental demand (M), physical demand (PD), temporal demand (TD), performance (P), effort (E), and frustration (F) are presented. $**$ stands for $p < 0.01$.}
\label{fig:results_2}
\vspace{-0.4cm}
\end{figure*}

In Fig. \ref{fig:results_2}, the boxplots of the NASA-TLX for all the subjects are reported in the two tested conditions. No statistically significant differences were observed across the evaluated metrics.

\vspace{-0.1cm}
\begin{table}[h]
\centering
\caption{Number of collisions with obstacles.}
\label{Tab:collision}
\begin{tabular}{llllll}
\toprule
\textbf{Controller} & \textbf{Corridor}     & $\mathbf{45^\circ}$ \textbf{L} & $\mathbf{45^\circ}$ \textbf{R} & \textbf{Front} & \textbf{Total}       \\
\midrule
AC  & 3 & 4 & 5 & 12 & 24\\
CB-MPC  & 1 & 1 & 0 & 0 & 2\\

\bottomrule \\ 
\multicolumn{3}{l}{}
\end{tabular}
\vspace{-10 mm}
\end{table}

\section{Discussion}
\label{sec: discussion}
Results of the \textit{structured navigation} experiments indicate that the CB-MPC approach yielded substantial improvements in both angular and linear energy consumption, with significant reductions compared to the AC. Importantly, no significant differences in mean jerk were observed between the two controllers. These results are further supported by the analysis of the subject’s mechanical work during the trials.  Consistent with the energy results, positive work was significantly reduced under the CB-MPC condition. As with the jerk metric, no significant differences in negative work were observed between the two controllers. Overall, these findings indicate that the improved energy efficiency achieved with CB-MPC did not come at the cost of increased oscillations and jerk.
In the \textit{unforeseen single-obstacle avoidance} experiment, the lack of significant differences in these metrics can be attributed to the shape and the reduced length of the path (approximately half of that in the first experiment), which limited the measurable differences in jerk and energy indicators, respectively. 

The primary objective of the \textit{unforeseen single-obstacle avoidance} experiment was to evaluate the safety level provided by the two controllers. By blindfolding the participants, visual feedback was removed, simulating a condition in which obstacle avoidance relied predominantly on the control strategy rather than on the user’s anticipatory and voluntary corrective capabilities. Under these conditions, the CB-MPC controller demonstrated a substantial improvement in safety during ambulation with I-WANDER in the presence of obstacles, significantly reducing collision events compared to AC. As shown in Table~\ref{Tab:collision}, when the obstacle was positioned at $45^{\circ}$, the AC controller successfully assisted the user in avoiding it in some trials. However, when the obstacle was placed directly in front of the robot, none of the participants avoided a collision under the AC condition. In contrast, as also illustrated in Fig.~\ref{fig:exp set up}, the CB-MPC controller enabled users to consistently avoid the obstacle without collisions, regardless of its placement. This proves the proposed MPC framework capacity to enhance safety through proactive obstacle avoidance. However, two collisions were also observed with the CB-MPC method. This may be due to the slack variable introduced to relax condition Eq. \ref{eq:cbf_slack_final} and preserve feasibility of the optimization problem. Such relaxation was required because obstacles were modeled as individual points, leading to a large number of constraints in the OCP and increasing computational complexity. As discussed below, this represents a limitation of the proposed framework, and addressing it in future work could further improve the safety of the method.

Although the quantitative performance metrics revealed significant differences between the controllers, no significant differences were observed in the NASA-TLX questionnaire scores. The corrective actions introduced by the CB-MPC to reduce collision risk enhance safety but may slightly diminish the perception of fully unconstrained, user-driven motion experienced during free walking in obstacle-free environments. For healthy individuals, who typically exhibit stable gait and good motor control, such safety-driven interventions may be perceived as limiting autonomy rather than supportive, potentially affecting usability. Conversely, for individuals with gait impairments, a strong sense of safety may be crucial to fostering confidence and facilitating safe ambulation. Achieving an effective balance between perceived freedom of movement and safety guarantees, therefore, becomes essential, and this balance may vary across users depending on factors such as the underlying pathology, its severity, psychological aspects, and other individual characteristics.

Although the second experiment simulated a condition of sensory impairment, a major limitation of the present study is the cohort of participants, which was exclusively healthy. To further validate the proposed method, especially from a usability and user-experience perspective, future studies will involve individuals with gait impairments (e.g., Parkinson’s disease or ataxia). This will allow us to assess the effectiveness and acceptance of the proposed control framework within the actual target population of I-WANDER, while enhancing demographic diversity and evaluating the robustness of the method in realistic clinical scenarios. Additional limitations concern the obstacle detection strategy adopted in this study. Indeed, obstacles were modeled as individual points, even when belonging to the same physical object. This requires considering a large number of obstacle points to ensure a sufficient representation of nearby objects, thereby increasing the number of constraints in the MPC formulation. A more structured approach that clusters obstacle points into unified objects with defined geometries would reduce the number of constraints to be enforced and lower the computational burden of the optimization problem. 
Furthermore, the estimation of the human’s intended direction of motion could be enhanced by integrating data-driven obstacle-aware prediction methods. Incorporating such intention forecasting into the OCP may further reduce energy consumption and jerk, leading to even smoother and more efficient assisted walking.

\section{Conclusion}
\label{sec: conclusions}
This work presented a control framework to enhance the safety of ambulation with walking assistive devices. The proposed approach was experimentally validated on the omnidirectional I-WANDER platform. A CBF formulation was embedded within an MPC framework to reduce the risk of collisions with obstacles while ensuring energy-efficient, smooth human–robot interaction. 
Although some limitations remain, including the exclusive involvement of healthy participants and a simplified point-based obstacle representation, the experimental results support the effectiveness of the proposed framework. Future work will extend validation to individuals with gait impairments and investigate improved obstacle modeling and intention prediction, paving the way for safer and more adaptive assistive walking systems.





\bibliographystyle{IEEEtran}
\bibliography{references.bib} 

\addtolength{\textheight}{-12cm}   

\end{document}